\documentclass[conference, letterpaper]{IEEEtran}
\usepackage{subcaption}
\usepackage{ifthen}

\newboolean{isfinalversion}
\setboolean{isfinalversion}{true}

\newcommand{\draft}[1]{\ifthenelse{\boolean{isfinalversion}}{\PackageError{draft}{Remove draft comment}{}}{#1}}
\newcommand{\levelBicing}[1]{\emph{#1}}

\usepackage[table]{xcolor}

\usepackage[cmex10]{amsmath}
\usepackage{color}
\usepackage{flushend}
\usepackage{fancyhdr}
\usepackage{hyperref}

\ifCLASSINFOpdf
   \usepackage[pdftex]{graphicx}
   \graphicspath{{images/}{jpeg/}}
  \DeclareGraphicsExtensions{.pdf,.jpeg,.png}
\else
\fi

\usepackage{url}

\renewcommand{\thispagestyle}[2]{}

\fancypagestyle{plain}{
        \fancyhead{}
        \fancyhead[C]{first page center header}
        \fancyfoot{}
        \fancyfoot[C]{first page center footer}
}
\begin{document}
%
\title{Predicting Occupancy Trends in Barcelona's Bicycle Service Stations 
Using Open Data}


\author{\IEEEauthorblockN{Gabriel Martins Dias, Boris Bellalta and Simon 
Oechsner}
\IEEEauthorblockA{~\\Department of Information and Communication Technologies\\ 
Universitat Pompeu Fabra, Barcelona, Spain\\
Email: \{gabriel.martins, boris.bellalta, simon.oechsner\}@upf.edu}
}

\maketitle

\begin{abstract}

In 2008, the CEO of the company that manages and maintains the 
public bicycle service in Barcelona\footnote{ 
\url{http://www.bicing.cat}} recognized that one may not expect to 
always find a place to leave the rented bike nearby their destination, 
similarly to the case when, driving a car, people may not find a parking 
lot\footnote{ 
\url{http://www.lavanguardia.com/vida/20080714/53501157289/}}.
In this work, we make predictions about the statuses of the stations of the public 
bicycle service in Barcelona. We show that it is feasible to correctly predict nearly 
half of the times when the stations are either completely full of bikes or completely 
empty, up to $2$ days before they actually happen.
That is, users might avoid stations at times when they could not return a bicycle that 
they have rented before, or when they would not find a bike to rent.
To achieve that, we apply the Random Forest algorithm to classify the status of 
the stations and improve the lifetime of the models using publicly available 
data, such as information about the weather forecast.
Finally, we expect that the results of the predictions can be used to improve 
the quality of the service and make it more reliable for the users.

\end{abstract}


\IEEEpeerreviewmaketitle

\section{Introduction}

The bicycle sharing system of Barcelona (called \emph{Bicing}) has more than 
$400$ stations and about $6.000$ bicycles that can be rented by users. It is a 
very common option for local people that prefer to protect the 
environment, look for an alternative to the traditional public transportation, 
want to exploit the benefits of bicycles or simply cannot afford to drive.

As any other system, nonetheless, it has some problems which users must deal 
with. 
One of the problems that most of the people notice in the first days of use 
is that the number of bicycles and the number of free slots are limited. That 
is, in some occasions it is not possible to find bikes near their location, and 
it may happen that they do not find an available slot at the stations which are 
close to their destination.
As one may notice, the problem of not finding a bike can be solved by taking 
another public transport, such as a bus, the subway or a train. However, 
if a person is currently using a bike and cannot leave it close to their 
destination, it may become an inconvenience and a reason for avoiding the use of 
the system in the future.

Under these circumstances, a prediction about how many bikes they will find in 
the next day at a certain station may improve the system's reliability and increase its usage. For 
example, if a person is planning to go to a place in the next morning, they 
must change their time of departure if there will be no bikes at the stations 
which are close to their origin, or if there will be no space to return their 
bike in a station near to their destination.

From the point of view of the service operator, they must build a schedule for 
their workers to collect bikes from a station which has many bikes to a station 
where only few bikes can be found at a certain time. This often happens, for 
example, at the stations placed in more elevated places, because the users take 
the bikes in the morning and drive downhill to their destinations, but they do not 
take them back, due to the difficulties ascending the streets.
Therefore, having a prediction about at which time and which stations will be 
completely full of bikes or completely empty, makes it possible to improve 
the quality of the service for the end users, besides better utilizing the 
resources and reducing their costs.

\begin{figure}[b]
	\centering
	\includegraphics[width=0.44\textwidth]{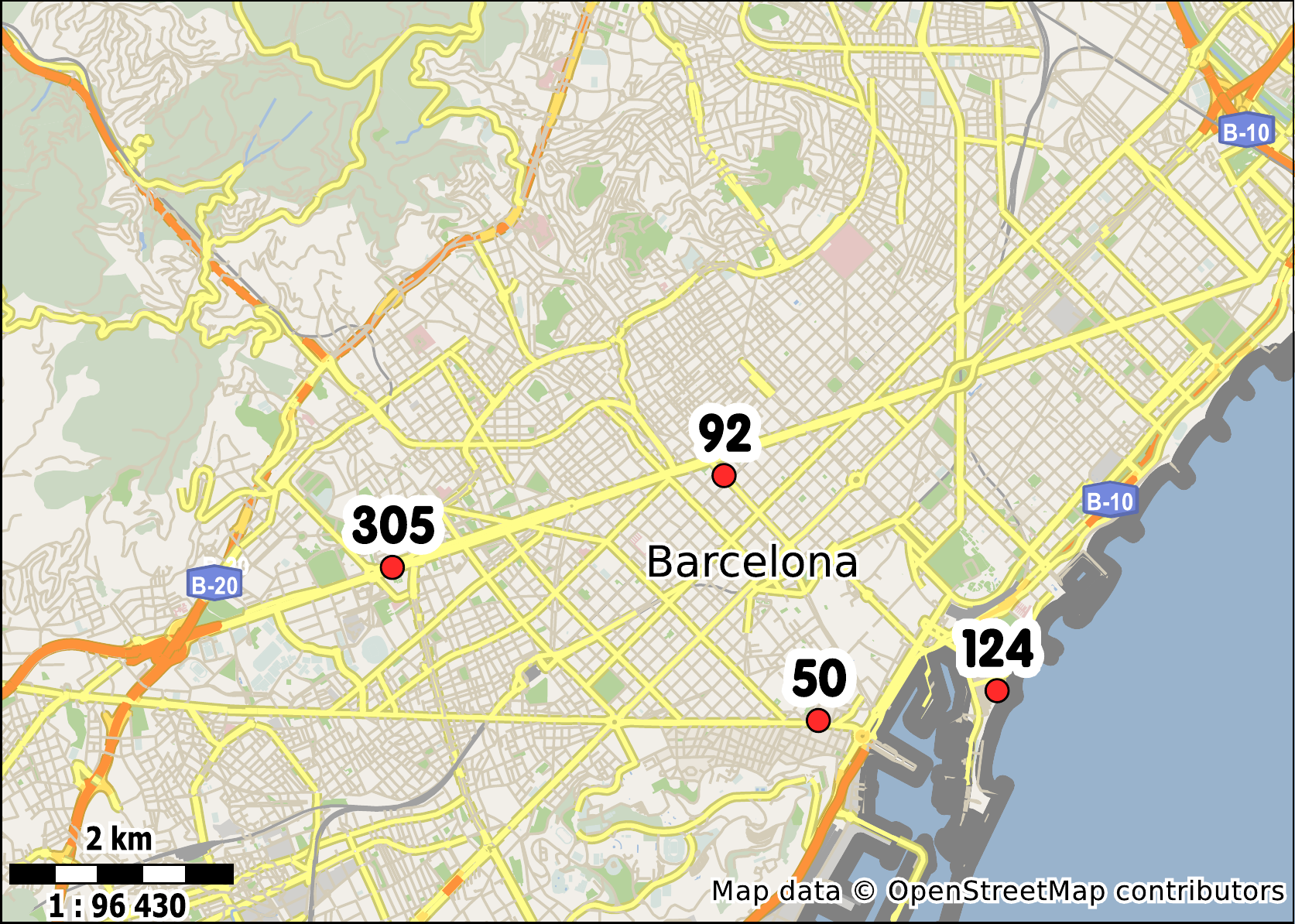}
	\caption{Location of the studied stations in Barcelona.}
	\label{fig:scenario}
\end{figure}

In this work, we focus on predict how many bicycles will be available at the stations 
at different times of the day. To achieve that, we apply the Random Forest algorithm 
and classify the observations according to the context of the city, i.e., 
considering that external factors (such as holidays, extremely low 
temperatures and rainfalls) may influence the behavior of the users and change the number of bikes rented on a specific day.
We utilize public data about the status of the stations, 
weather and festivities (available online) and predict the statuses of the stations 
up to $72$ hours earlier.

\begin{figure*}[t]
        \centering
        \begin{subfigure}[t]{0.22\textwidth}
                \centering
                \includegraphics[width=\textwidth]{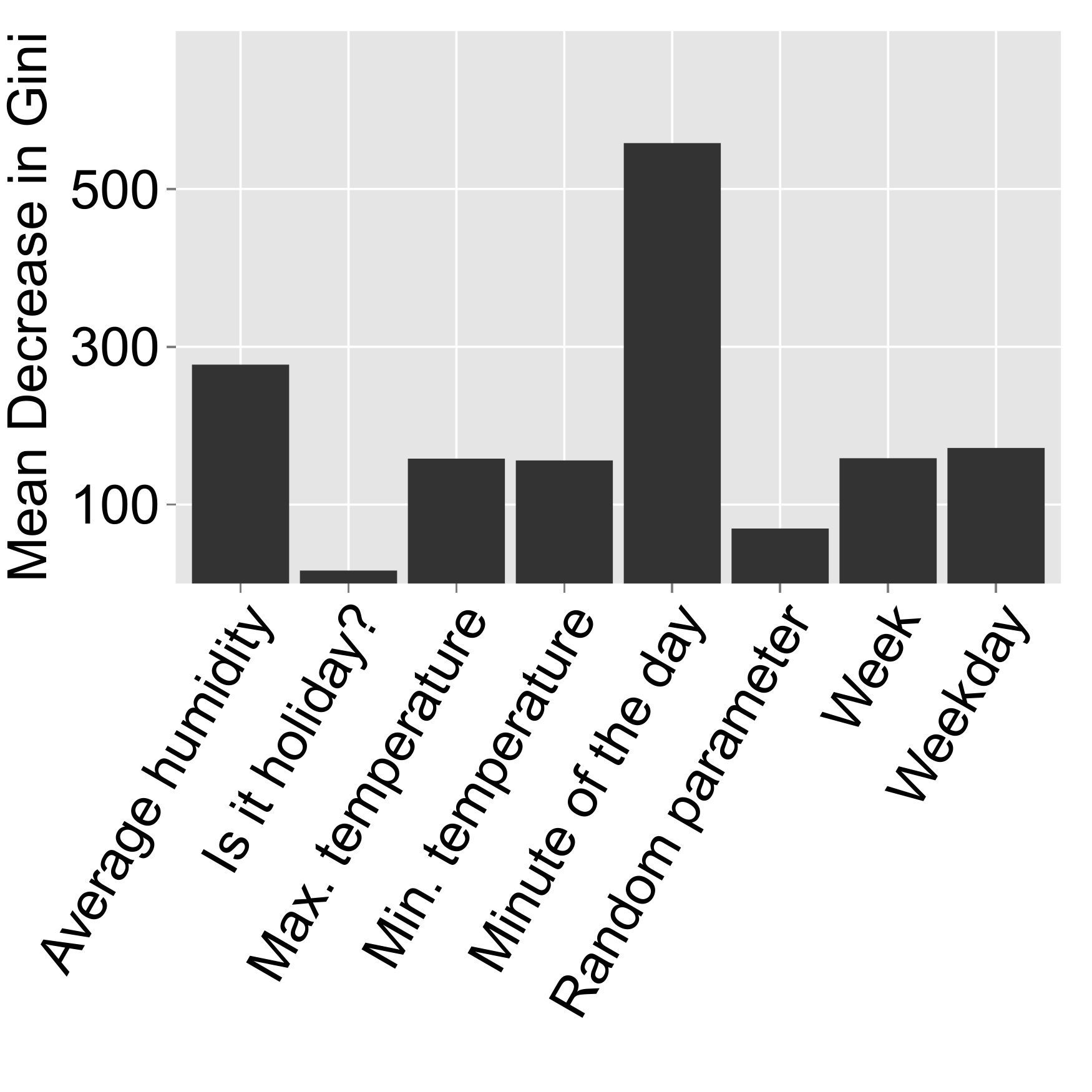}
                \caption{Station 50}
                \label{fig:importance-50}
        \end{subfigure}%
        \qquad
        \begin{subfigure}[t]{0.22\textwidth}
                \centering
                \includegraphics[width=\textwidth]{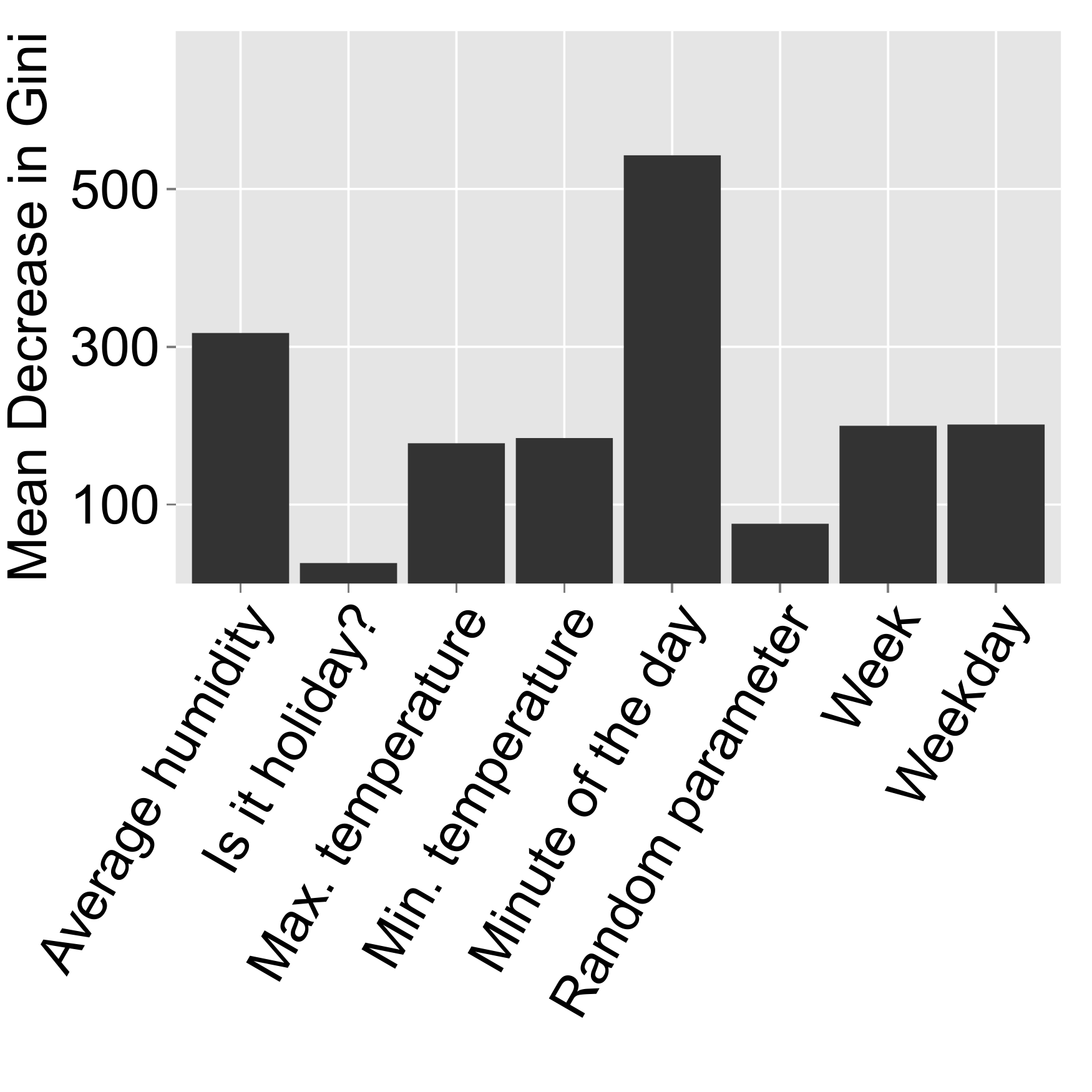}
                \caption{Station 124}
                \label{fig:importance-124}
        \end{subfigure}%
        \qquad
         \begin{subfigure}[t]{0.22\textwidth}
                \centering
                \includegraphics[width=\textwidth]{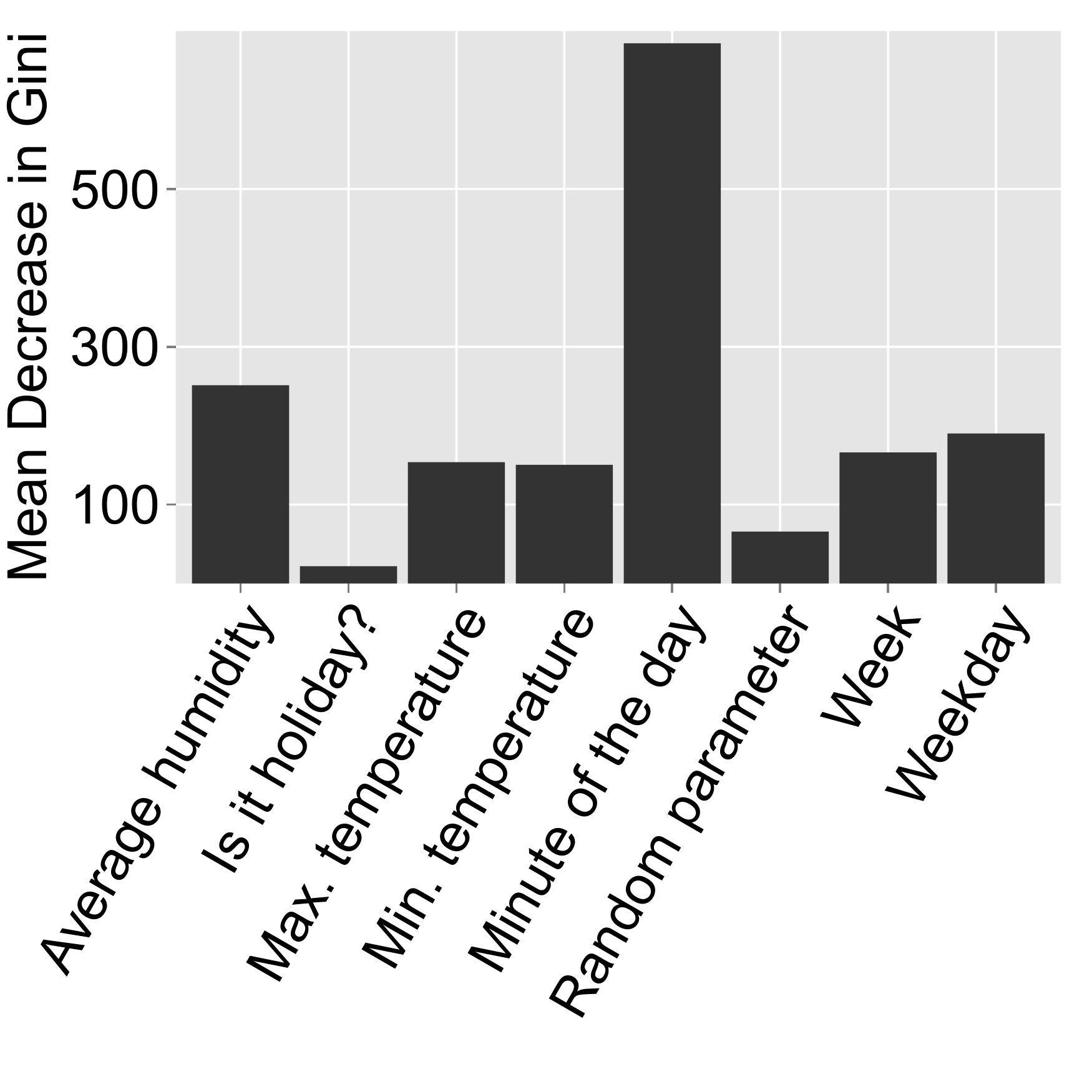}
                \caption{Station 92}
                \label{fig:importance-92}
        \end{subfigure}%
	\qquad
        \begin{subfigure}[t]{0.22\textwidth}
                \centering
                \includegraphics[width=\textwidth]{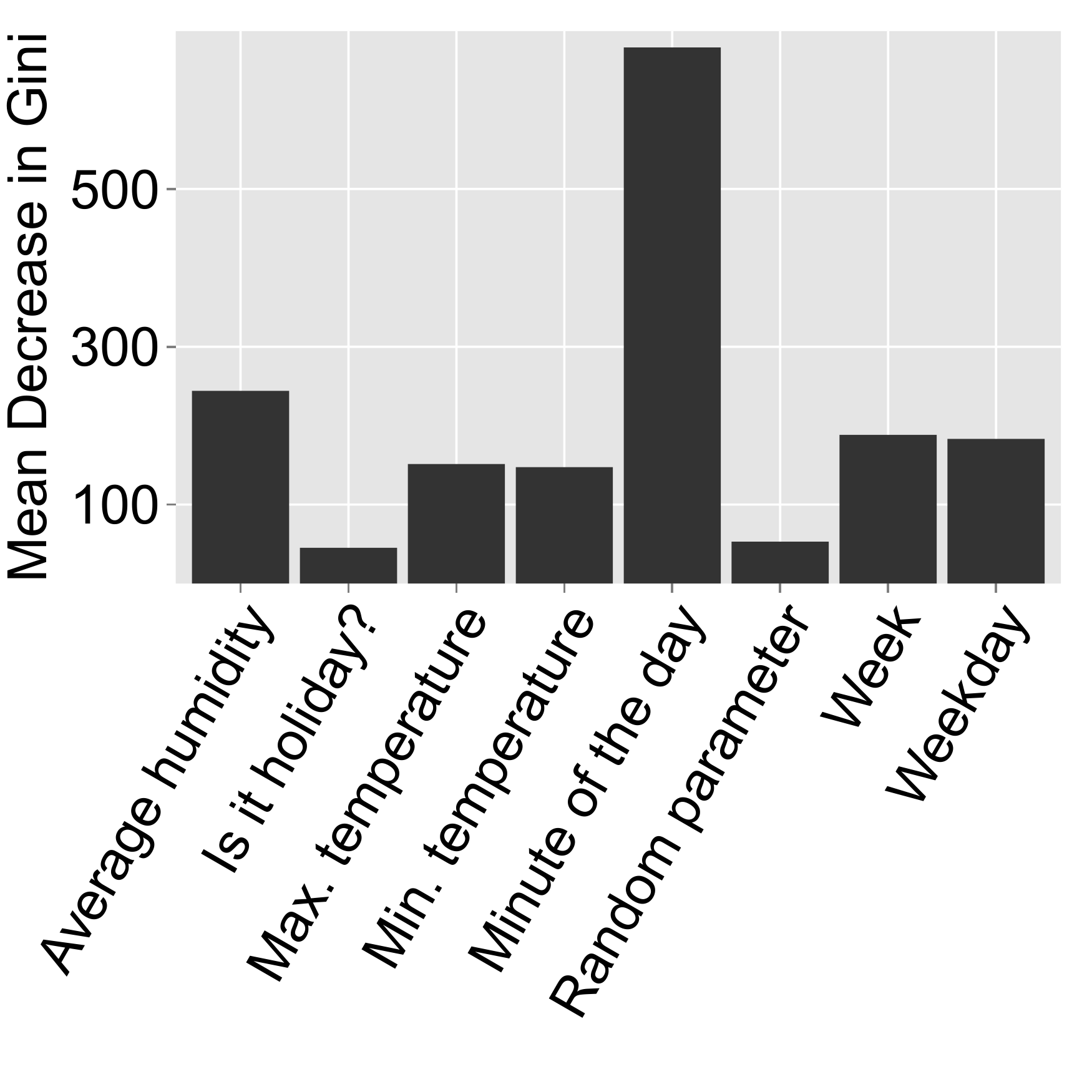}
                \caption{Station 305}
                \label{fig:importance-305}
        \end{subfigure}%
        \caption{Importance of the external factors in the number of bikes at 
the observed stations.}
        \label{fig:importance}
\end{figure*}

Our contribution is a way to make predictions about the status of the 
bike stations using several data types.
We illustrate the performance of the predictions and compare them to other 
approaches that do not provide as accurate responses as those achieved by the 
Random Forest models.
Moreover, we show the potential of exploring public data to improve large-scale 
systems, considering that this kind of predictions may be also applied to other 
cities that use similar systems, or to different public services, such as the 
public parking system.

In Section~\ref{section:background}, we explain the related work and give an 
overview on prediction methods.
Later, in Section~\ref{section:data-sets}, we explain the data sets used and 
how to access them, before showing how we utilized the data in 
Section~\ref{section:study-of-the-data}.
In Section~\ref{section:results}, we show and explain the performance of the 
predictions done during the tests.
Finally, we draw some conclusions about the methods used and possible 
applications for them in Section~\ref{section:conclusion}.


%

\section{Background}
\label{section:background}

In this Section, we give an overview about how predictions are made and list the 
most relevant works that use different prediction methods based on public data.

\subsection{Time Series}

There are many methods developed to predict values from sampled data.
Na\"ive predictions are based on aggregate functions, e.g., the average, the 
sum, the maximum or the median value among a set of numbers. 
Although they have a low computational cost, their accuracy is usually 
constrained.

Time series methods are more powerful. They consider the evolution of a system, 
given historical observations indexed by time. Examples of this kind of method 
may vary from the simplest Autoregressive (AR) and Moving Average (MA) to the 
most complex extensions of ARIMA (SARIMA, VARIMA, ARIMA-GARCH)\cite{Box1990}. 
Such methods are more often employed in economics in order to predict 
stock prices and their variations, but their applications can be extended to 
almost any area of study.
In comparison with the na\"ive methods, they consume more computational power, 
but the expectations about their accuracy are also much higher. 
In~\cite{Kaltenbrunner2010}, the authors did an analysis of the \emph{Bicing} 
system and made predictions about the number of bikes in the near future using 
Na\"ive and AR models. In their experiments, they showed that predicting the 
number of bikes in the next hour using AR models was much more accurate than 
assuming that they would be the same as in the last observations.
Time series methods are widely used to predict one-dimensional data. However, in 
complex environments--such as a city-- there exist a high number of parameters 
which may have an impact on the system that is going to be studied.

\subsection{Multivariate methods}

In traditional prediction methods that use only time series as input, the 
data is considered univariate. That is, apart from the time (used to index the
observations), only the own variable (represented by past observations) is 
used as input to predict the future values.
Considering that the input variables are used to make predictions, we will call 
them \emph{predictors}.
In many cases, the data can be multivariate and a set of predictors will be 
mapped to the output value.
For example, besides having the time to index each observation, it may be possible 
to use as well characteristics of the environment that are relevant and may 
cause impact in the future output values.
As real world situations are usually composed by several aspects which may
impact the others, such methods may reproduce better the evidences of
the environment.
Corroborating such idea, recent studies have concluded that considering multiple 
sources of data and different points of view can produce more accurate 
predictions~\cite{Mellers2015}.
Therefore, there is a need for multivariate prediction methods, which are able to 
incorporate several sources of information in order to produce more accurate
results.

In~\cite{Froehlich2009}, the authors studied the bike system operation with 
predictions about the number of bikes at the stations in the near future using 
Na\"ive models and Bayesian Networks. Before making the predictions, it was 
necessary to classify every station according to the average level of use per 
day, because each prediction is made for a certain group. 
Additionally, every time a new station is installed, its usage must be compared with 
the usage of the other stations in order to verify which of them have similar levels.
The main restriction of their approach is that it ignores the impact of external 
factors, such as a large event in a public space, which may attract more users 
to the closest bike stations.
That is, in order to change the current models, a new data analysis would be 
required to include these parameters, before generating a new model in a 
procedure with a space and time complexity which may increase exponentially. 
In our approach, new predictors can be added without requiring extra 
computation apart from the creation of the models.

In~\cite{Lathia2010}, the authors observed the similarity of the subway stations 
from a different perspective than ours. They described a method to build clusters of  
stations based on the data collected from users behavior and proposed a mechanism 
to make predictions based on such knowledge.
Finally, they proposed three prediction methods that are based on weighted average
between similar trips.
As before, the main limitation of this method is that the computational 
costs for including new parameters are extremely high--if possible.
That is, during the system development, it is necessary to determine what 
the most relevant events are that may affect the use of every single station and 
select those that show a high correlation, based on historical observations.


As an example of a scalable multivariate method, decision tree learning~\cite{Quinlan1986} 
is successfully applied in machine learning problems to make predictions.
A decision tree is a n-ary tree that has a height equal to the number of 
predictors plus one, where the last level contains the leaves and represents 
the output values.
The general idea of using a decision tree as a predictive model, 
is to create a way to systematically map the characteristics of an observation 
(composed by several variables) to its output value.
If the output value is a label instead of a number, the decision tree may be 
called a classification tree.

\subsection{Random Forest}

Random Forest~\cite{Breiman2001} is an algorithm that uses decision trees 
to create unbiased classification trees.
To achieve that, the procedure randomly selects a subset of predictors from the 
original data and builds a classification tree based on it.
After creating several trees, a final decision tree is built based on the 
average relevance of the predictors, and can be used to make 
predictions over other sets of observations.
Therefore, besides building the decision tree, the Random Forest algorithm  
calculates the relevance of each predictor in the real environment.

One method to represent the relevance of the predictors is the calculation of 
the mean decrease in the Gini coefficient of the predictors.
The Gini coefficient is a measure of inequality of a distribution, where 0 
represents perfect equality and 1 perfect inequality. 
In the Random Forest algorithm, after the creation of the decision trees, the 
Gini coefficient is calculated for each level of the tree and compared to their 
respective parents.
Changes in the Gini coefficients are summed for each predictor 
in all trees and normalized in order to calculate their mean decrease.
In summary, the mean decrease in Gini coefficient is a measure of how each 
predictor contributes to the homogeneity of the nodes and leaves in the 
resulting decision tree. 
A higher decrease in Gini coefficient of one predictor means that such predictor 
is more relevant when classifying an observation.

In this work, we adopted the Random Forest as the prediction method. From the 
best of our knowledge, it is the first study that uses Random Forest to 
make predictions over publicly available data. Furthermore, it can also produce 
better results for longer time intervals and larger data sets, as well as 
providing the scalability lacked in the other works.









\section{Data sets}
\label{section:data-sets}

In this Section we are going to explain how the data used in the experiments 
is handled. This description makes this research reproducible, besides explaining 
the source and the reliability of the data which we used in the tests.

The data generated by the \emph{Bicing} system is open for public access via an 
API (Application Programming Interface) that can be accessed 
online\footnote{\url{http://wservice.viabicing.cat/v1/getstations.php?v=1}}. 
The information available in this data set is about the current status of the 
stations: their location; a list of the closest stations; whether they 
are operating normally or not; the number of bikes available; and the number of 
free slots (i.e., the number of bikes that can be parked at that station at 
that moment). This information is updated almost once a minute.
For more than one year, we stored this information nearly once a minute, i.e., 
despite some occasional failures, such as connectivity problems between our 
computers and our database server,
we have all the information generated by the system in this time interval 
available for the tests.

Besides the information about the bike system, we accessed two other sources of 
data that contain information about factors that may affect the operation of \emph{Bicing}: 
the holidays and the weather (including the forecast) in the city.
The data about the holidays in 
Barcelona\footnote{\url{http://www.bcn.cat/calendarifestius/en/}} have been 
chosen because they involve the national and local festivities, such as bank 
and religious holidays, and we expect that on these days the normal activities 
in shops, banks, schools and universities, are suspended or reduced.
Moreover, the weather 
information\footnote{
\url{http://www.wunderground.com/history/airport/LEBL/2015/}}, such as 
temperature, relative humidity, dew point and wind speed, may be correlated to 
changes in the users' behavior and influence the use of the bikes.
Finally, the weather forecast provides detailed information about the current 
local weather and how it is expected to be in the next 3 days. That is, it 
incorporates the predictions about the weather that usually contain a 
combination of several factors and are computed by supercomputers which are not 
accessible by every person.

In conclusion, the external information is mainly chosen according to its descriptive 
power about the situation of the city in the past, besides providing a 
prediction about whether such scenarios will occur again in the future.


\section{Study of the Data}
\label{section:study-of-the-data}

In this Section we explain how we transformed the raw data, which is 
available online for open access, into analyzable information. For example, the 
procedures to clean and analyze the data before performing tests and applying 
the results.

\begin{table}[h]
\centering
\def\arraystretch{1.5}
\rowcolors{2}{gray!25}{white}
\begin{tabular}{
	>{\centering\arraybackslash}p{2cm}
	>{\centering\arraybackslash}p{4cm}}
\rowcolor{gray!50}
\textbf{Level} &
\textbf{Description} 
 \\ 
\hline 
Full & No slots available \\ 
\hline 
Almost full & Between 1 and 5 slots available \\ 
\hline 
Bikes and slots available & More than 5 bikes available and more than 5 slots 
available \\ 
\hline 
Almost empty & Between 1 and 5 bikes available \\ 
\hline 
Empty & No bikes available \\ 
\hline 
\end{tabular} 
\caption{Levels of bikes and slots available}
\label{table:levels}
\end{table}

First of all, we phrase our problem from the point of view of a normal user. 
Since they cannot rent more than one bike at a time, it is not required to have 
a prediction about the exact number of bikes at a certain time in the future. 
It is more relevant, on the other hand, to know whether there will be bikes or not at 
such a moment.
Therefore, we classify the status of a station as \levelBicing{full}, \levelBicing{almost 
full}, \levelBicing{slots and bikes available}, \levelBicing{almost empty} or \levelBicing{empty}, as 
shown in Table~\ref{table:levels}.

\begin{figure}[t]
	\centering
	\includegraphics[width=0.33\textwidth]{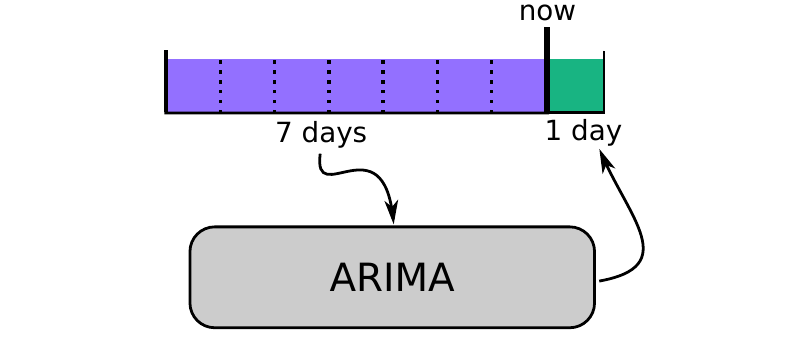}
	\caption{The data selected for the predictions using ARIMA are from
	the last $7$ days before the model creation.}
	\label{fig:rf-predictors-arima}
\end{figure}

The data set with the information from the bike system is large 
(around 10 Gigabytes).
Thus, in order to make predictions using an ARIMA model, we selected the 
last $7$ days (approximately $10.000$ observations), as shown in 
Figure~\ref{fig:rf-predictors-arima}.
The number of available bikes at the stations was considered a time series and 
a seasonal ARIMA model was chosen using the Akaike information 
criterion~\cite{Akaike1974}.
For the seasonal factor, we assumed that every day has a similar usage pattern 
and the statuses tend to repeat through the days.
In the end, we classified the predictions according to the levels explained 
before (\levelBicing{full}, \levelBicing{almost full}, etc.).

\begin{figure}[t]
	\centering
	\includegraphics[width=0.33\textwidth]{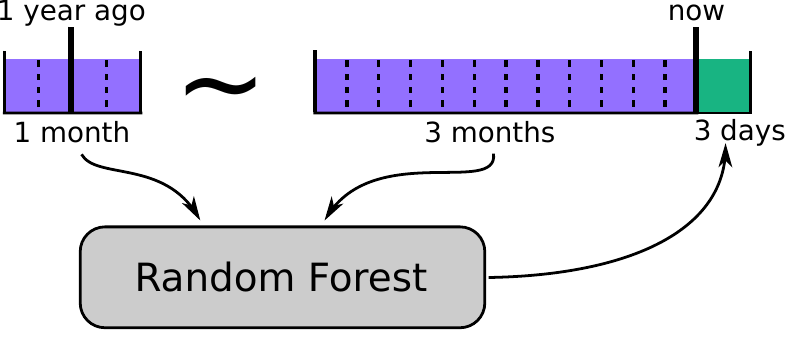}
	\caption{For the predictions using Random Forest, the data from
	the last $3$ months were merged with the data centered $1$ year before
	the model creation.}
	\label{fig:rf-predictors}
\end{figure}

The second type of predictions was done using the Random Forest algorithm.
A subset of the data was created by randomly selecting $1.000$ entries from 
the period that starts $54$ weeks before the model creation and finishes $50$ 
weeks before it, i.e., a total of $4$ weeks. Additionally, we randomly select $2.000$ entries from a period 
that starts $3$ months before the model creation.
Figure~\ref{fig:rf-predictors} shows which data were selected in order to generate
the predictions about the next days.

We select this dataset for two reasons, the first one is due to the memory 
required to construct a model, which is nearly $7$ Gigabytes when we use
$3.000$ entries. Given that we cannot select all available data, we try to
choose observations that are (probably) the most similar to the next
days, which are going to be predicted.
Having built this data set, we are able to construct two different sets of 
predictors for the models, one that uses only the data extracted from the bike 
system (RF) and another one that merges them with the data from the weather forecast 
and the information about holidays (Extended RF, explained in Section~\ref{section:data-sets}).

Given that there is no space to show the status of every station, we 
observed the data from the last week of January $2015$ and selected $4$ 
stations which were either completely full or completely empty during more than 
30\% of the time: stations number $50$, $124$, $92$ and $305$.
In Figure~\ref{fig:scenario}, it is possible to see a map of Barcelona and their exact 
position. Each station is at least $2$ kilometers from the others and has 
distinct characteristics. For example, station $124$ is very close to the 
sea, while station $305$ is at $60$ meters above the sea level; stations 
$50$ and $92$ are very close to subway stations and the others are not.

Based on the mean decrease of Gini coefficient, we can observe in 
Figure~\ref{fig:importance} how external factors correlate with the 
statuses of the stations during the whole year. 
For example, it is possible to notice that the most relevant factor for all of 
them is the time of the day, i.e., their daily use is comparable. 
For example, if we observed that, at station $92$, there were more bikes on 
Monday morning than on Monday afternoon, we might expect that on Tuesday morning 
there would be more bikes than on Tuesday afternoon.
This illustrates that our assumption about the inter-day seasonality (when 
using ARIMA models) is valid.

Although this is relevant and explored in other studies that observed daily 
trends~\cite{Froehlich2009}, there are interesting points which have not 
been noticed before. For example, at station $124$, the relative humidity 
has a greater effect than at the others, which can be explained by its 
location (close to the sea, where people tend to avoid going when it is raining). 
At stations $50$ and $92$, the day of the week is more relevant than at the 
others, which can also be explained by their location: since they are close to 
subway stations, they become an option for workers and students that use the 
public bikes in combination with the subway to arrive at their destination.
In conclusion, we can see that each station has its own "profile", i.e., some 
external factors influence more the use of some stations than of the others.

\section{Results}
\label{section:results}

\begin{figure*}[t]
        \centering
        \begin{subfigure}[t]{0.22\textwidth}
                \centering
                \includegraphics[width=\textwidth]{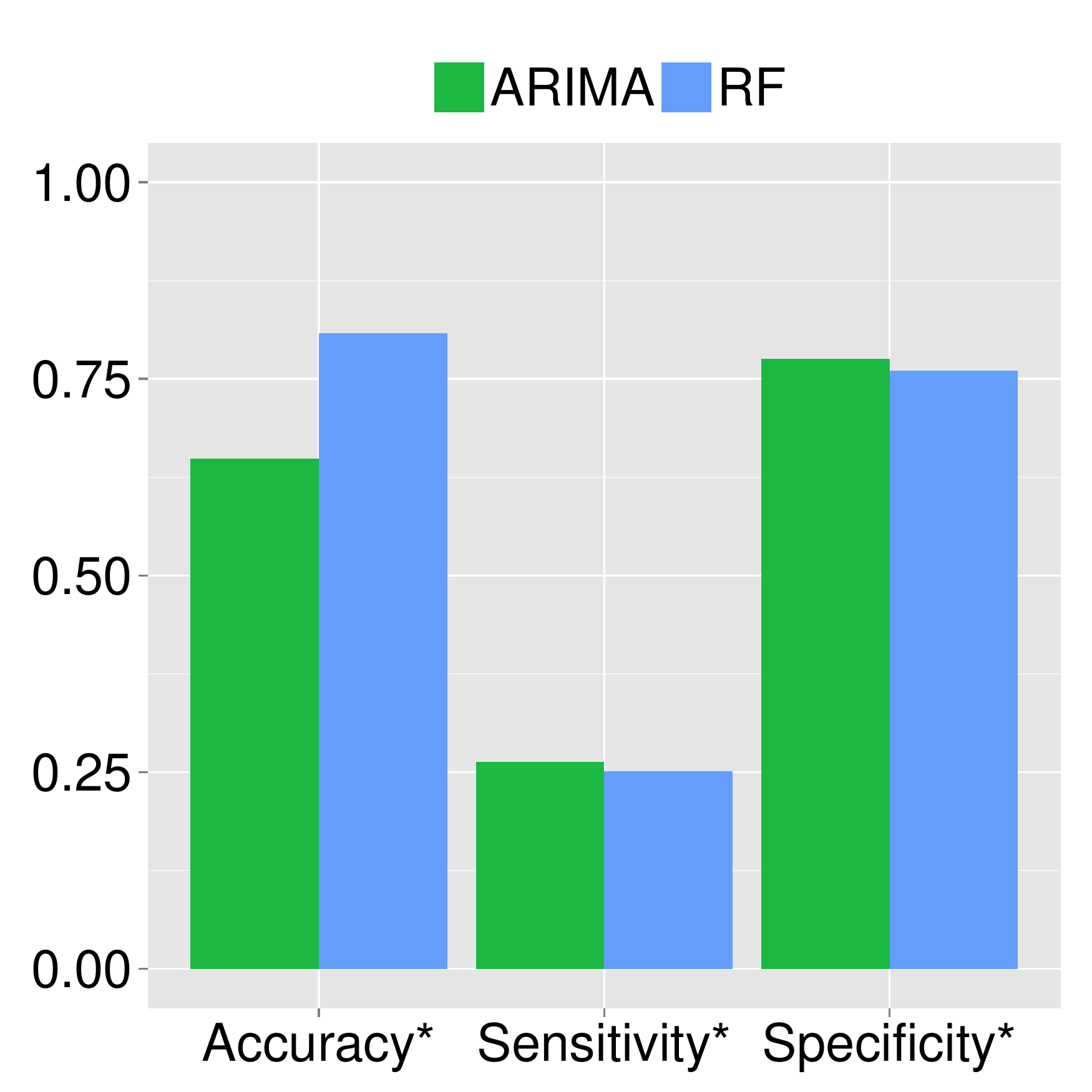}
                \caption{Station 50}
                \label{fig:arima-vs-rf-50}
        \end{subfigure}%
        \qquad
        \begin{subfigure}[t]{0.22\textwidth}
                \centering
                \includegraphics[width=\textwidth]{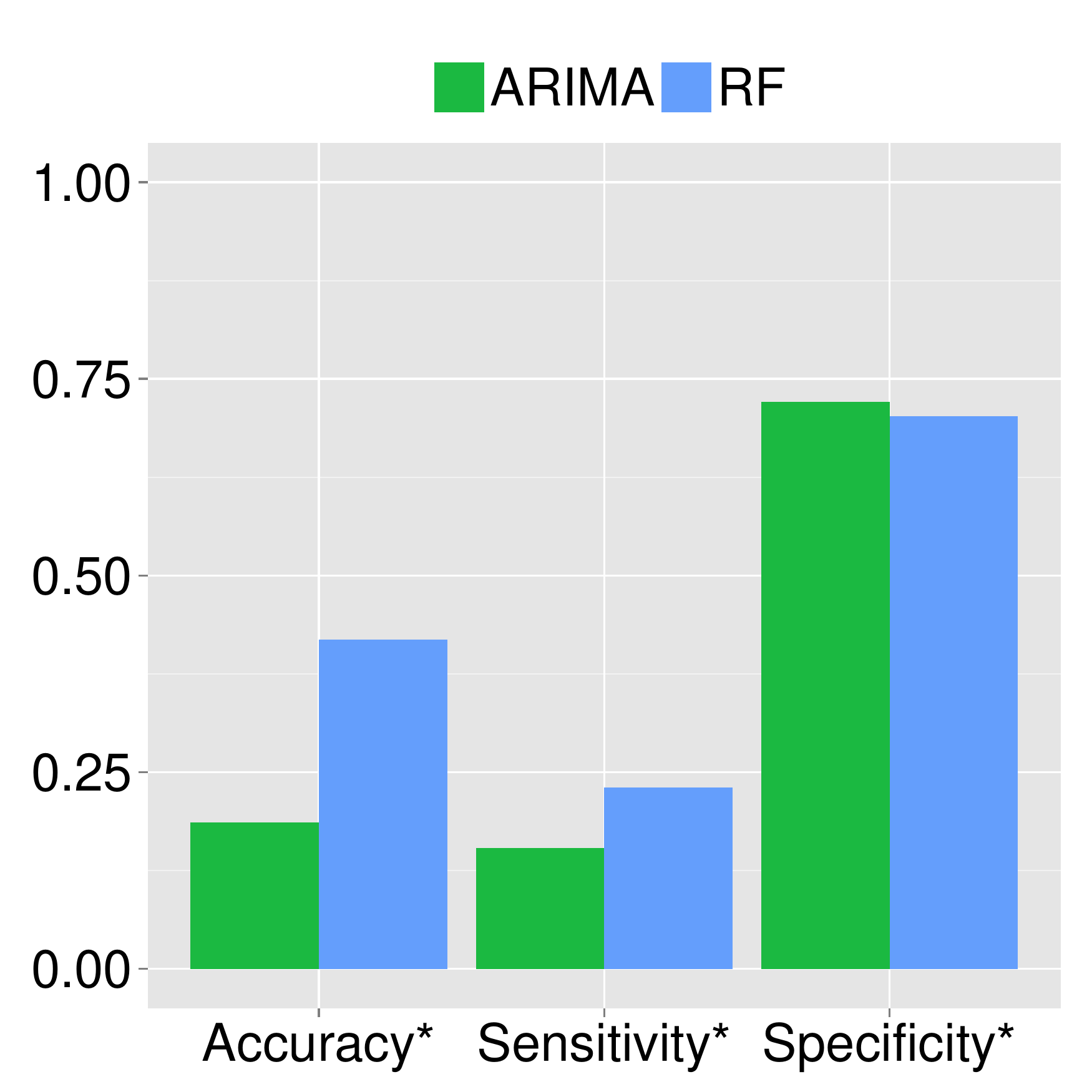}
                \caption{Station 124}
                \label{fig:arima-vs-rf-124}
        \end{subfigure}%
        \qquad
         \begin{subfigure}[t]{0.22\textwidth}
                \centering
                \includegraphics[width=\textwidth]{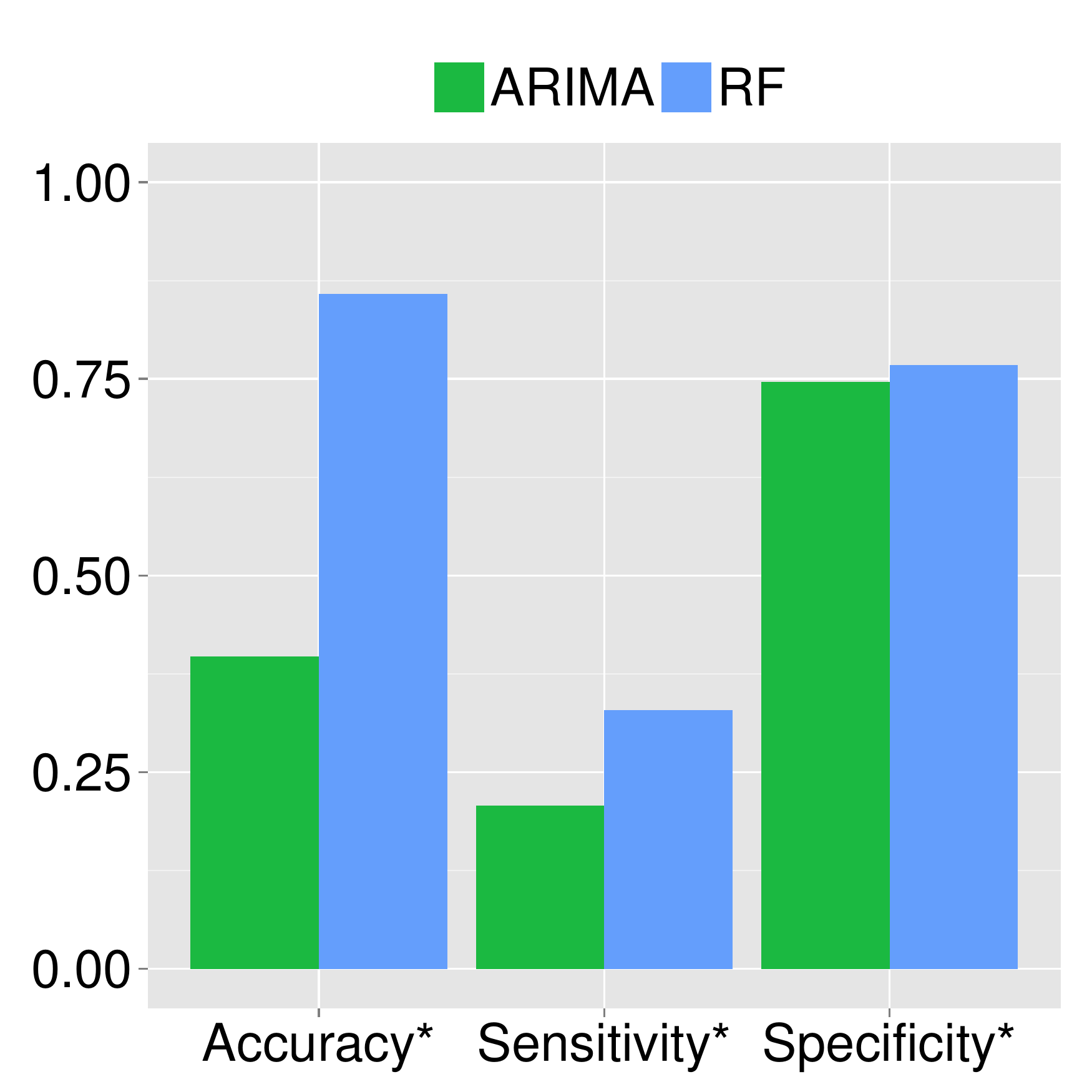}
                \caption{Station 92}
                \label{fig:arima-vs-rf-92}
        \end{subfigure}%
	\qquad
        \begin{subfigure}[t]{0.22\textwidth}
                \centering
                \includegraphics[width=\textwidth]{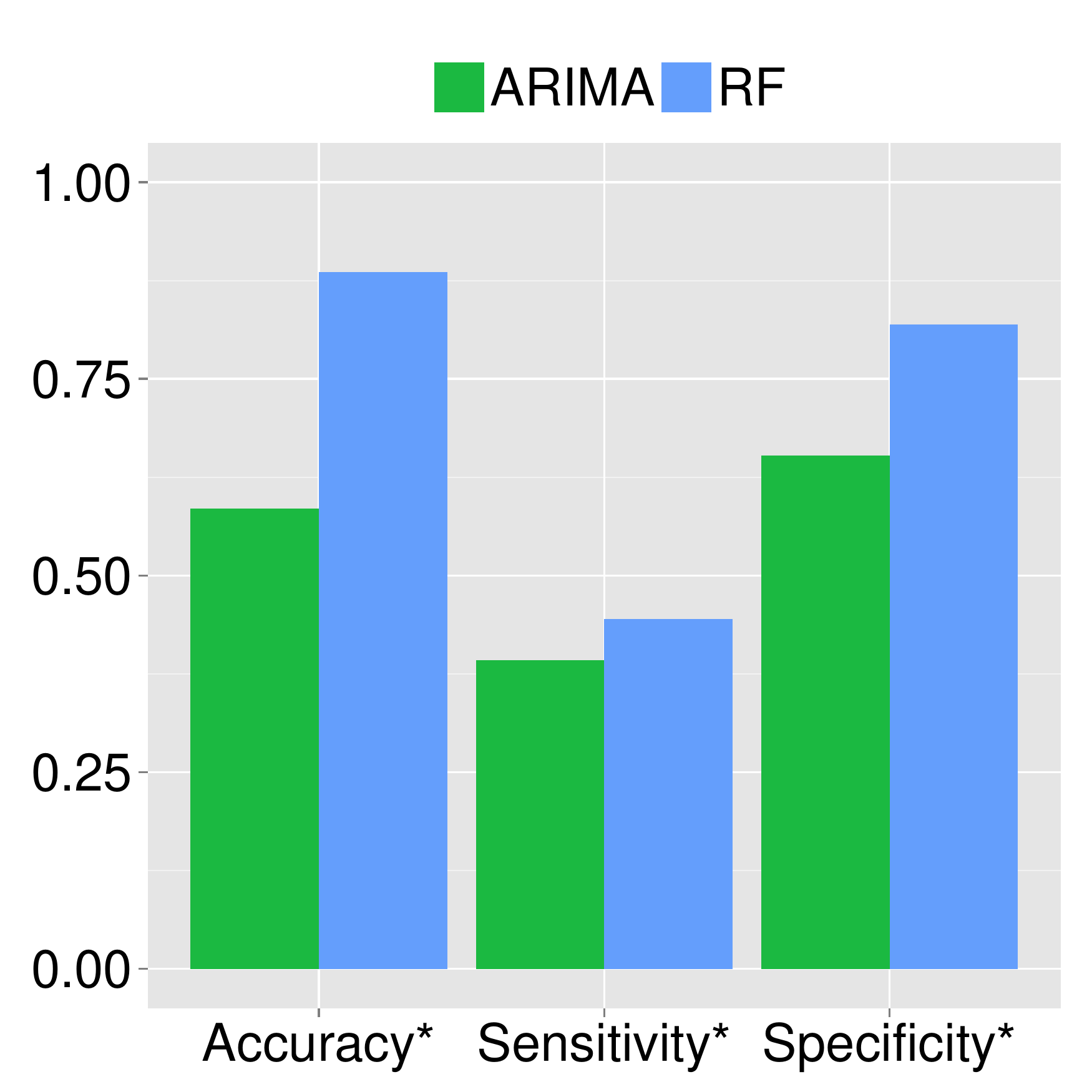}
                \caption{Station 305}
                \label{fig:arima-vs-rf-305}
        \end{subfigure}%
        \caption{Performance of the different methods using a more flexible 
criteria to evaluate.}
        \label{fig:arima-vs-rf}
\end{figure*}


During February 5--12, 2015, we ran our tests based on the selected 
stations. From 00:00 to 01:00 we set up the prediction models and made 
predictions for every $15$ minutes. 
We calculated the accuracy as the share of time when the stations were either 
\levelBicing{empty} and \levelBicing{full} and had been successfully predicted. 
Moreover, in order to observe the performance of the predictions, we calculated 
the sensitivity and the specificity of each method.

The \emph{sensitivity} is the relation between the total number of 
``true positives'' and the ``positives''.
In our scenario, a ``positive'' happens when a station is predicted to be either 
\levelBicing{full} or \levelBicing{empty}.
Thus, a ``true positive'' happens when the station was actually observed 
being \levelBicing{full} or \levelBicing{empty} after having made such a prediction. 
Therefore, if a prediction model has higher sensitivity,
it  means that the predictions about the "positives" are more reliable than with a model
with lower rates.

Analogously, the \emph{specificity} is the relation between the total number of 
``true negatives'' and the number of ``negatives''.
In our predictions, an outcome is considered as a ``negative'' when it predicts 
that the station will be neither \levelBicing{full} nor \levelBicing{empty} and 
a ``true negative'' happens if the station was neither \levelBicing{full} nor 
\levelBicing{empty} at that moment.
Again, if a prediction model has higher specificity, it  means that the predictions 
about the "negatives" are more reliable than using a model with lower rates.

\subsection{ARIMA vs. Random Forest}

Predictions using ARIMA models had been shown to be very efficient in the short term 
(less than 15 minutes) by other authors.
However, our tests showed that they are very inaccurate when predicting a complete day.
In fact, they had never predicted that the stations would be completely full 
nor completely empty, because they tend to be around an average from the last couple 
of days.
After observing such a low accuracy, we decided to include a more flexible evaluation 
of the results.
The flexible criteria (marked with an asterisk in the plots) considers 
predictions of \levelBicing{almost full} and \levelBicing{almost empty} levels 
as ``positive'' outcomes when calculating the sensitivity and specificity.
From a practical perspective, users and administrators can consider these levels 
as "warnings" about the possibility of achieving a critical level 
(\levelBicing{full} or \levelBicing{empty}).

In Figure~\ref{fig:arima-vs-rf} we show the obtained results for the 
predictions using ARIMA and those using the Random Forest models.
Even considering a more flexible model of evaluation, the ARIMA models
are significantly outperformed by the predictions obtained using the Random 
Forest algorithm. The biggest difference occurred at station $92$, 
where the ARIMA models correctly predicted $40\%$ of the critical statuses,
while the accuracy of the Random Forest algorithm was $86\%$.
The sensitivity of the predictions using the Random Forest algorithm is larger
for three stations and the difference for the fourth one (station $50$) was
$1\%$.
Finally, the specificity of the predictions using the Random Forest algorithm are 
similar for three stations, and the biggest difference is of about $16\%$ 
for station $305$.

\subsection{Enhanced Random Forest models}

\begin{figure*}[t]
        \centering
        \begin{subfigure}[t]{0.22\textwidth}
                \centering
                \includegraphics[width=\textwidth]{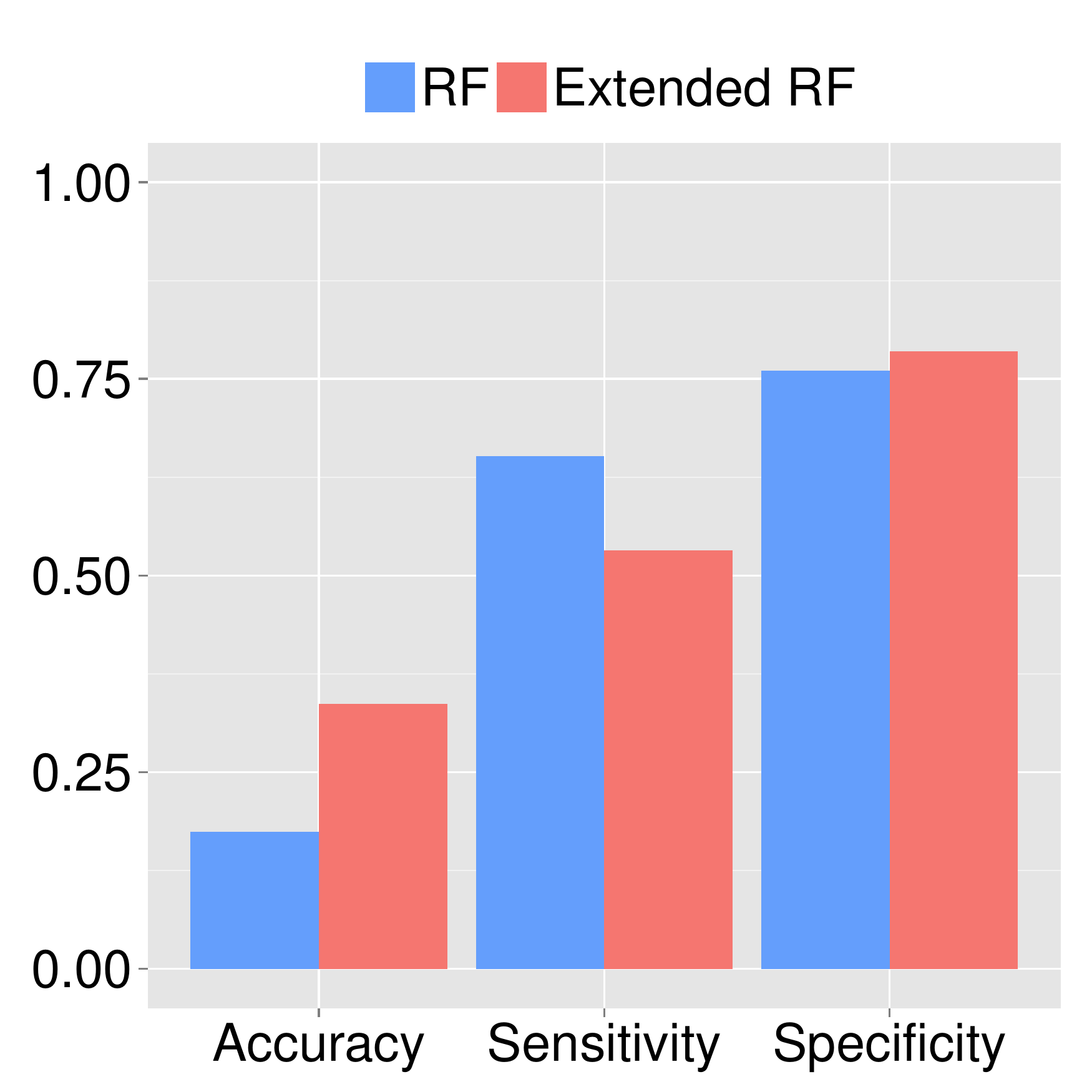}
                \caption{Station 50}
                \label{fig:rf-complete-50}
        \end{subfigure}%
        \qquad
        \begin{subfigure}[t]{0.22\textwidth}
                \centering
                \includegraphics[width=\textwidth]{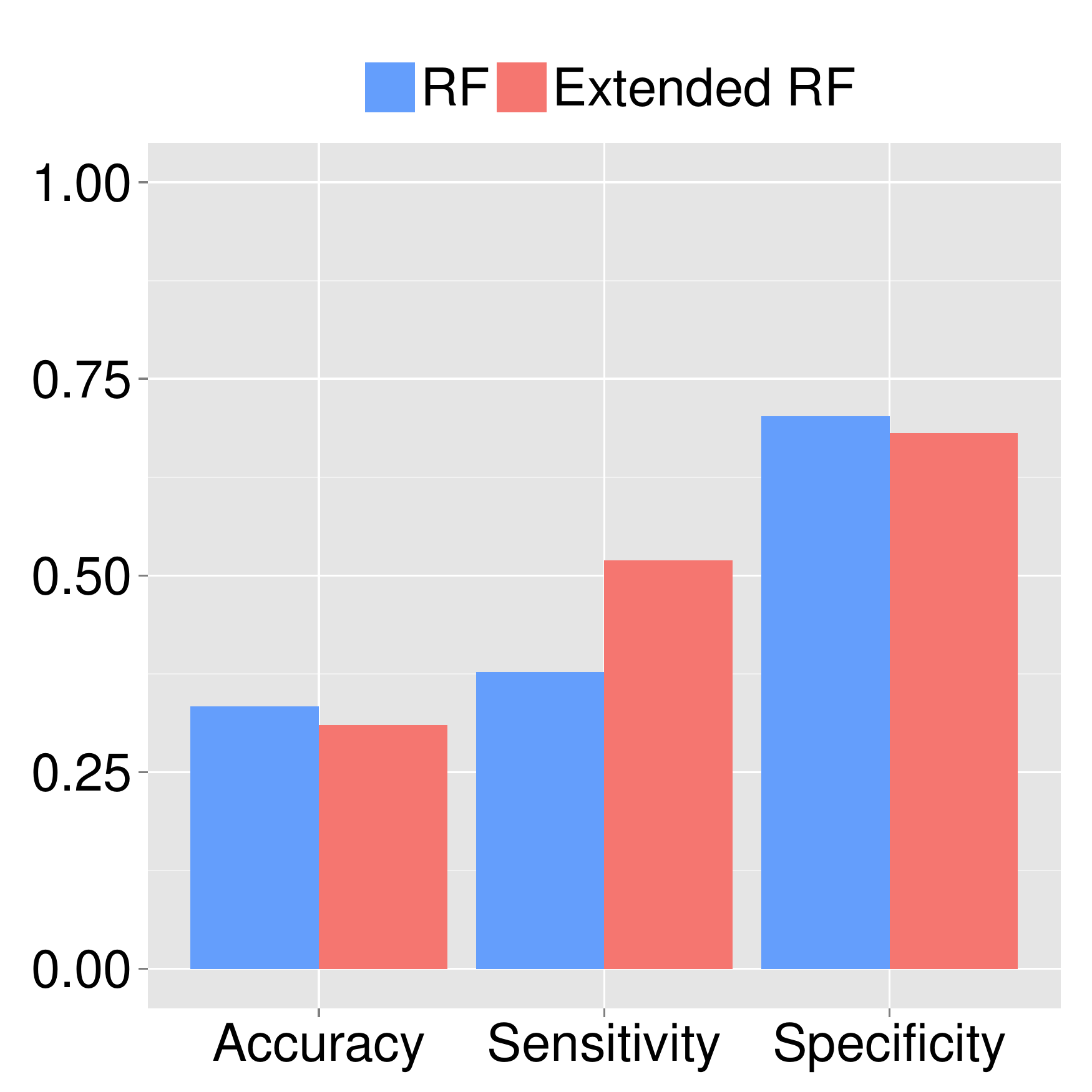}
                \caption{Station 124}
                \label{fig:rf-complete-124}
        \end{subfigure}%
        \qquad
         \begin{subfigure}[t]{0.22\textwidth}
                \centering
                \includegraphics[width=\textwidth]{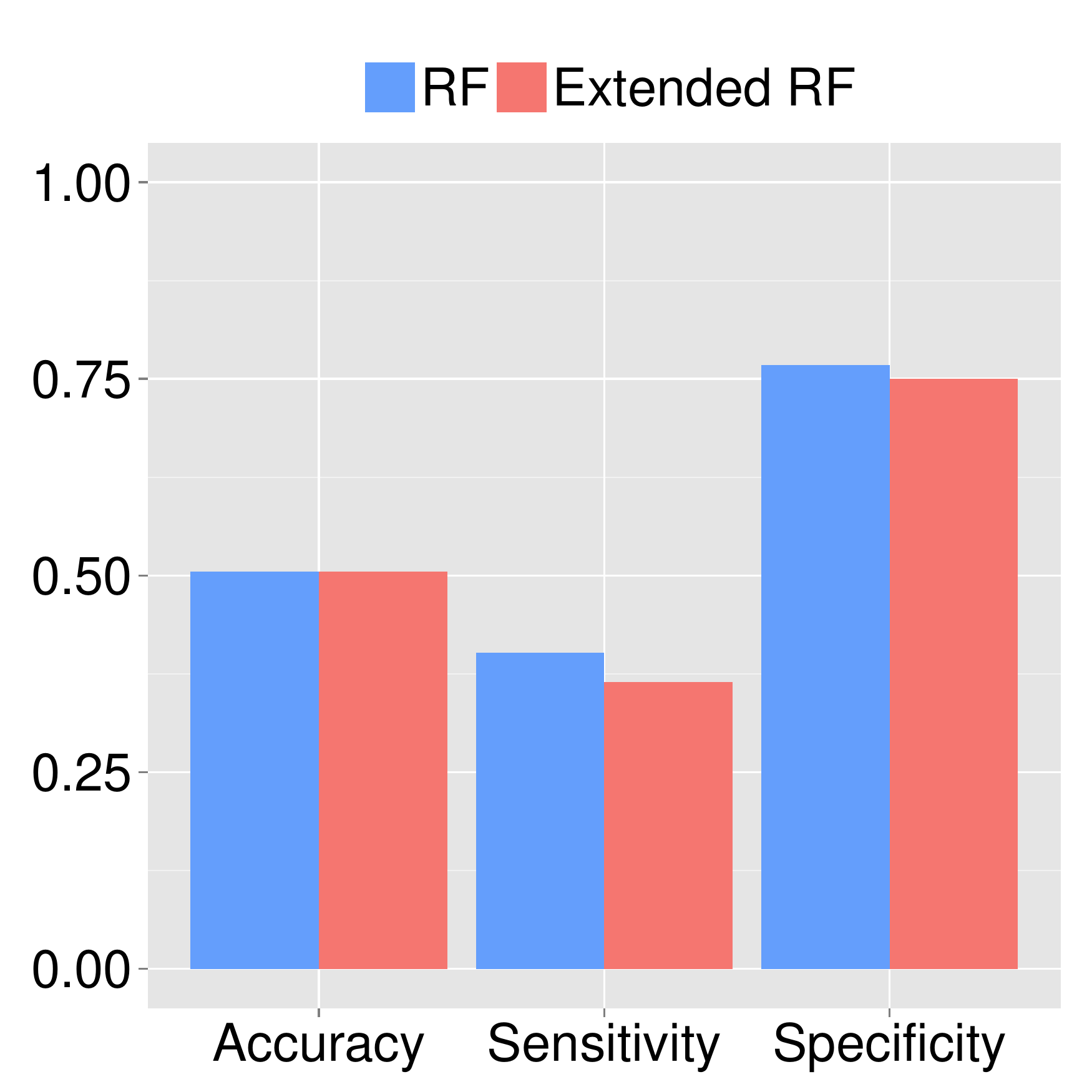}
                \caption{Station 92}
                \label{fig:rf-complete-92}
        \end{subfigure}%
	\qquad
        \begin{subfigure}[t]{0.22\textwidth}
                \centering
                \includegraphics[width=\textwidth]{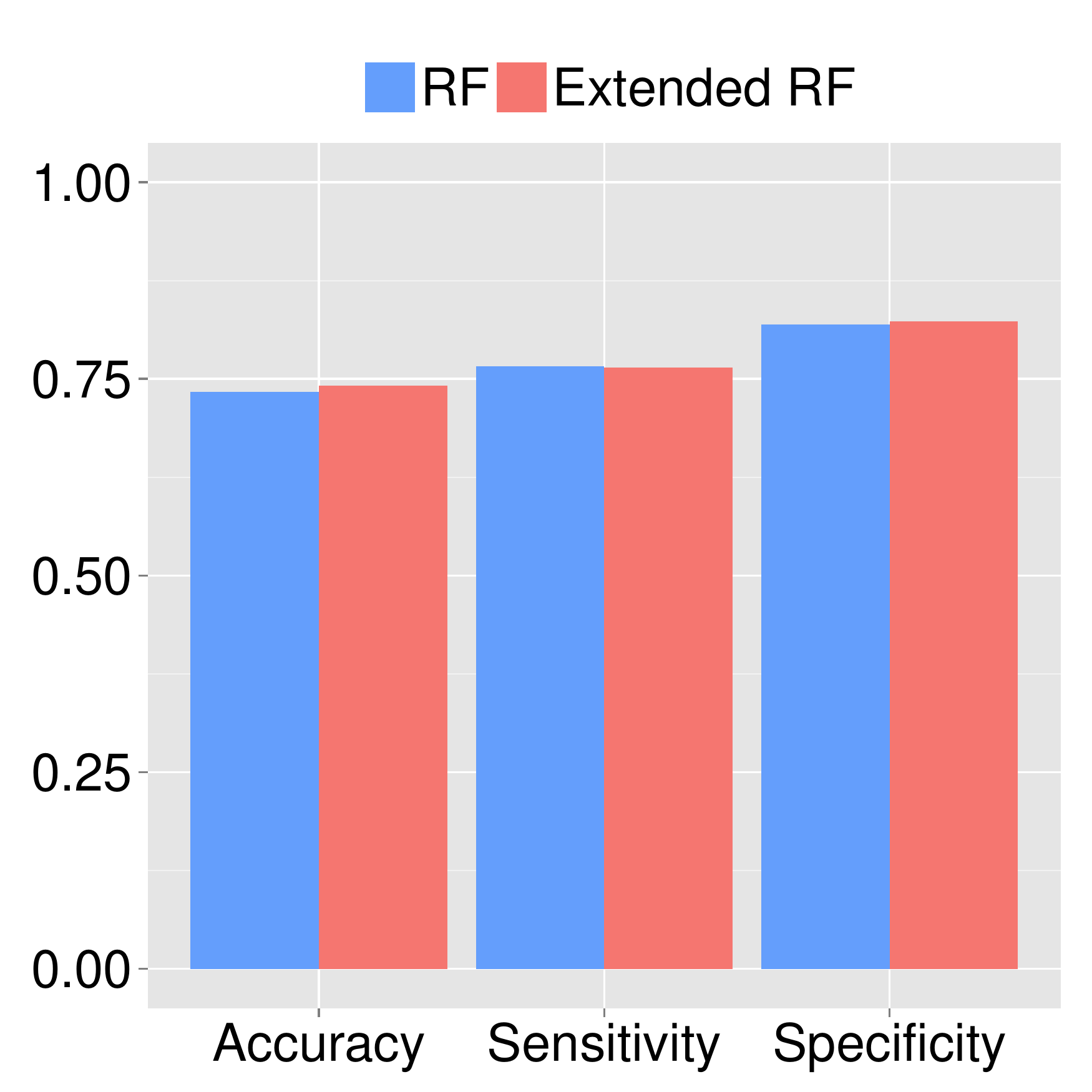}
                \caption{Station 305}
                \label{fig:rf-complete-305}
        \end{subfigure}%
        \caption{Performance of the Random Forest models in the next $24$ hours 
using different predictors.}
        \label{fig:rf-complete}
\end{figure*}

In Figure~\ref{fig:rf-complete} we compare two different Random Forest models, 
the ``RF'' and the ``Extended RF''. The difference is that the simplest Random Forest 
models (``RF'') have fewer predictors and, because of this, they require less 
data and less effort to be built.
However, the other models (``Extended RF'') incorporate the predictions 
about the weather conditions and the information about possible holidays, which 
may increase the amount of information that they consider when making a 
prediction.

Due to the low sensitivity observed using the flexible criteria, we included
only the results obtained based on the standard criteria of evaluation for 
their performance, which considers strictly the outcome of the predictions as 
their exact meaning, i.e., a prediction outcome is considered as a ``positive'' 
only if it is either \levelBicing{full} or \levelBicing{empty}, and as a 
``negative'' otherwise.
We notice that the sensitivity and the specificity of those 
predictions are always better than those obtained by the ARIMA models showed 
before.

Although there is no big difference for stations $92$ and $305$, 
for station $124$ the difference between the sensitivity of the predictions 
that use external information and the predictions that use only the data 
from the \emph{Bicing} system was around $15\%$ and for station $50$, 
it was $-12\%$.
On average, they were the same.

When comparing the different stations, with our results we can affirm that some 
of them are "less predictable" than the others. For example, for stations $124$ 
and $92$ the accuracy was lower than  $40\%$, while for station $305$ it was 
almost $75\%$.
Considering all the observed stations together, both methods had sensitivity of approximately $49\%$ 
and specificity of more than $74\%$.
As we can also observe, the differences were not relevant for most of the 
stations, and we conclude that there were no gains on including external 
information, considering the predictions made for the next $24$ hours.


\subsection{Durability of the models}

Considering that the high total number of stations may require a large computation 
time in order to create the prediction models and to make the predictions, we observed whether 
the quality of the predictions had decreased over several days.
That is, we made predictions for the next $3$ days ($72$ hours)
and compared their accuracy, sensitivity and specificity according 
to the age of the predictions about the same days (from 
February $7$ to $12$).

\begin{figure*}[t]
        \centering
        \begin{subfigure}[t]{0.22\textwidth}
                \centering
                \includegraphics[width=\textwidth]{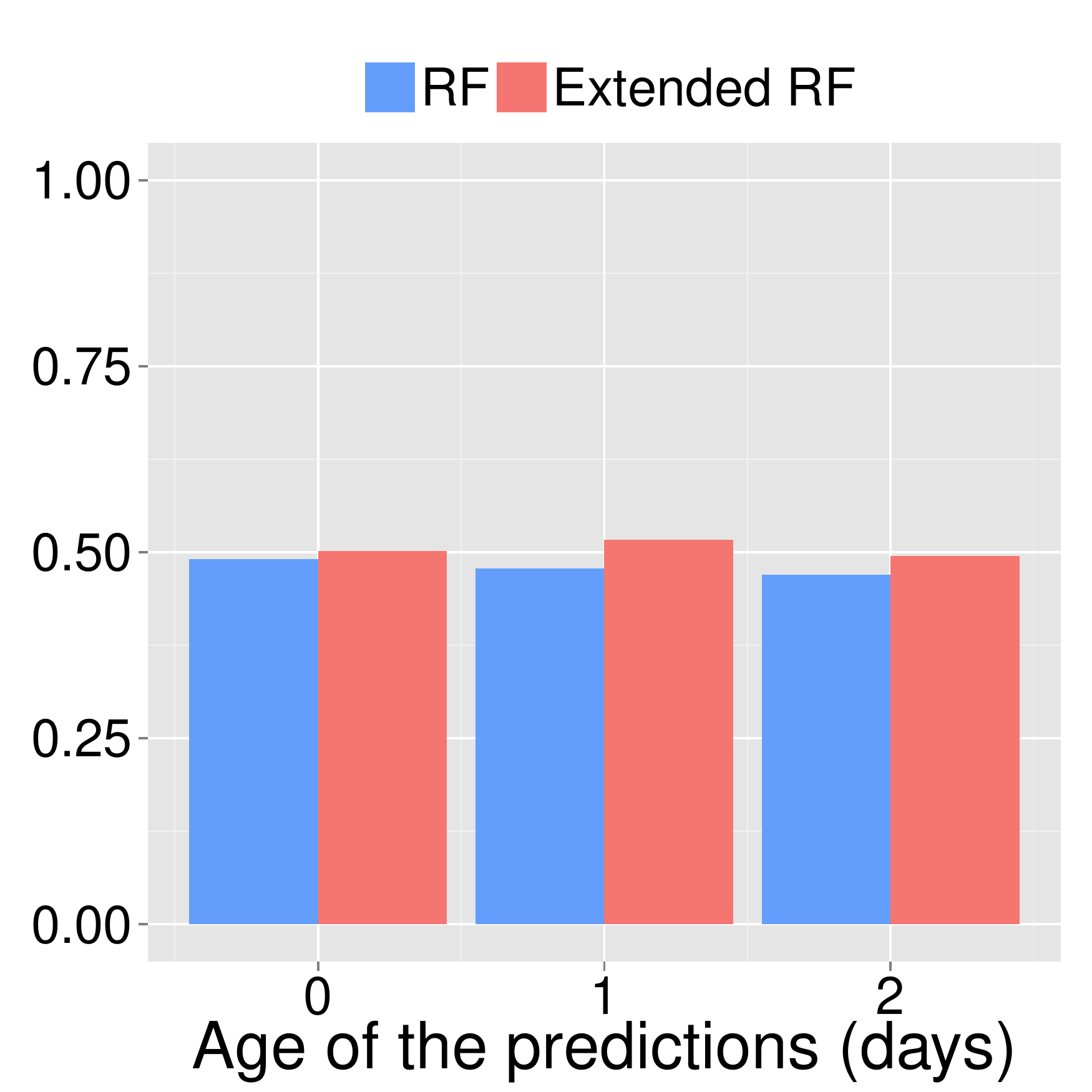}
                \caption{Accuracy}
                \label{fig:rf-consolided-Accuracy}
        \end{subfigure}%
        \qquad        
        \begin{subfigure}[t]{0.22\textwidth}
                \centering
                \includegraphics[width=\textwidth]{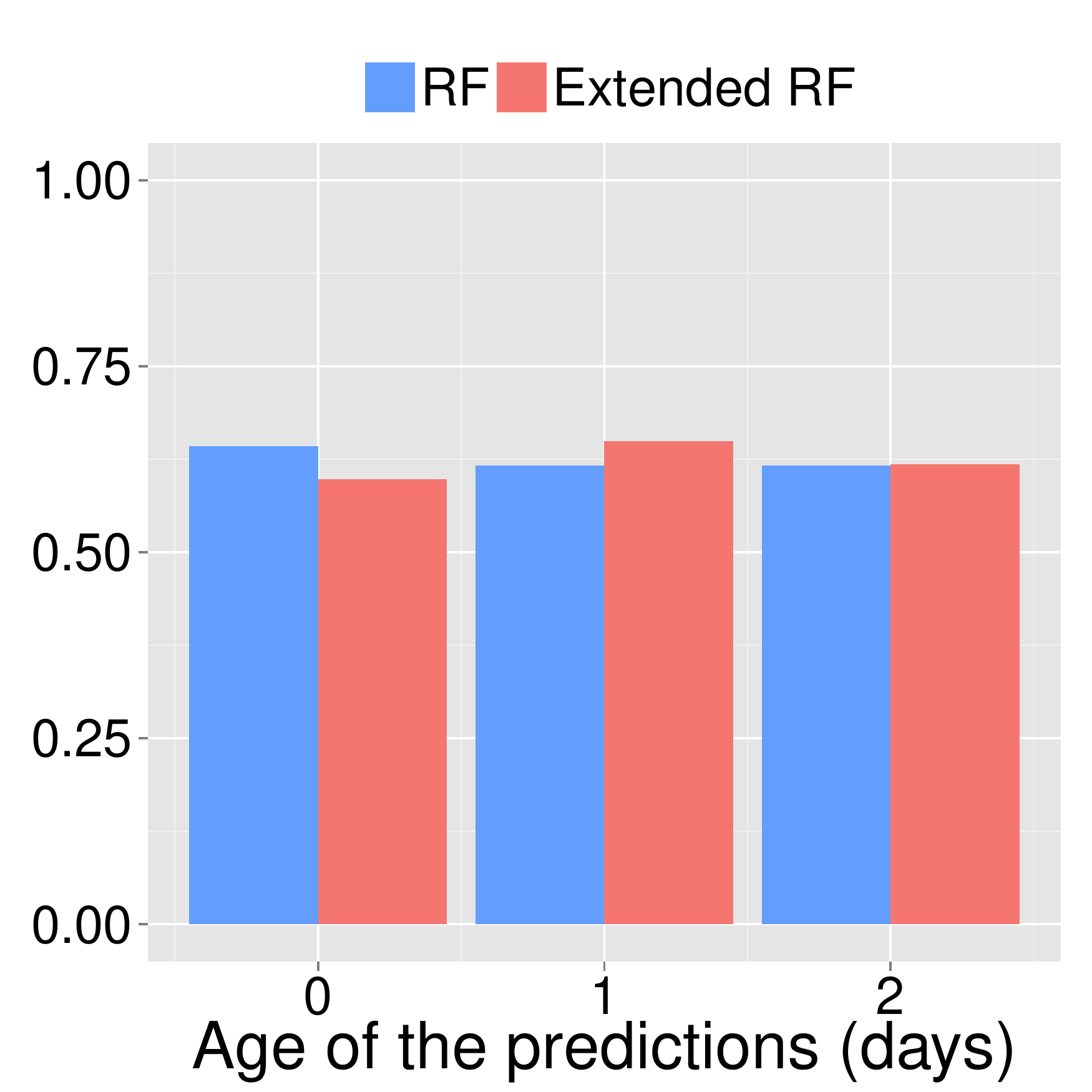}
                \caption{Sensitivity}
                \label{fig:rf-consolided-Sensitivity}
        \end{subfigure}%
        \qquad
        \begin{subfigure}[t]{0.22\textwidth}
                \centering
                \includegraphics[width=\textwidth]{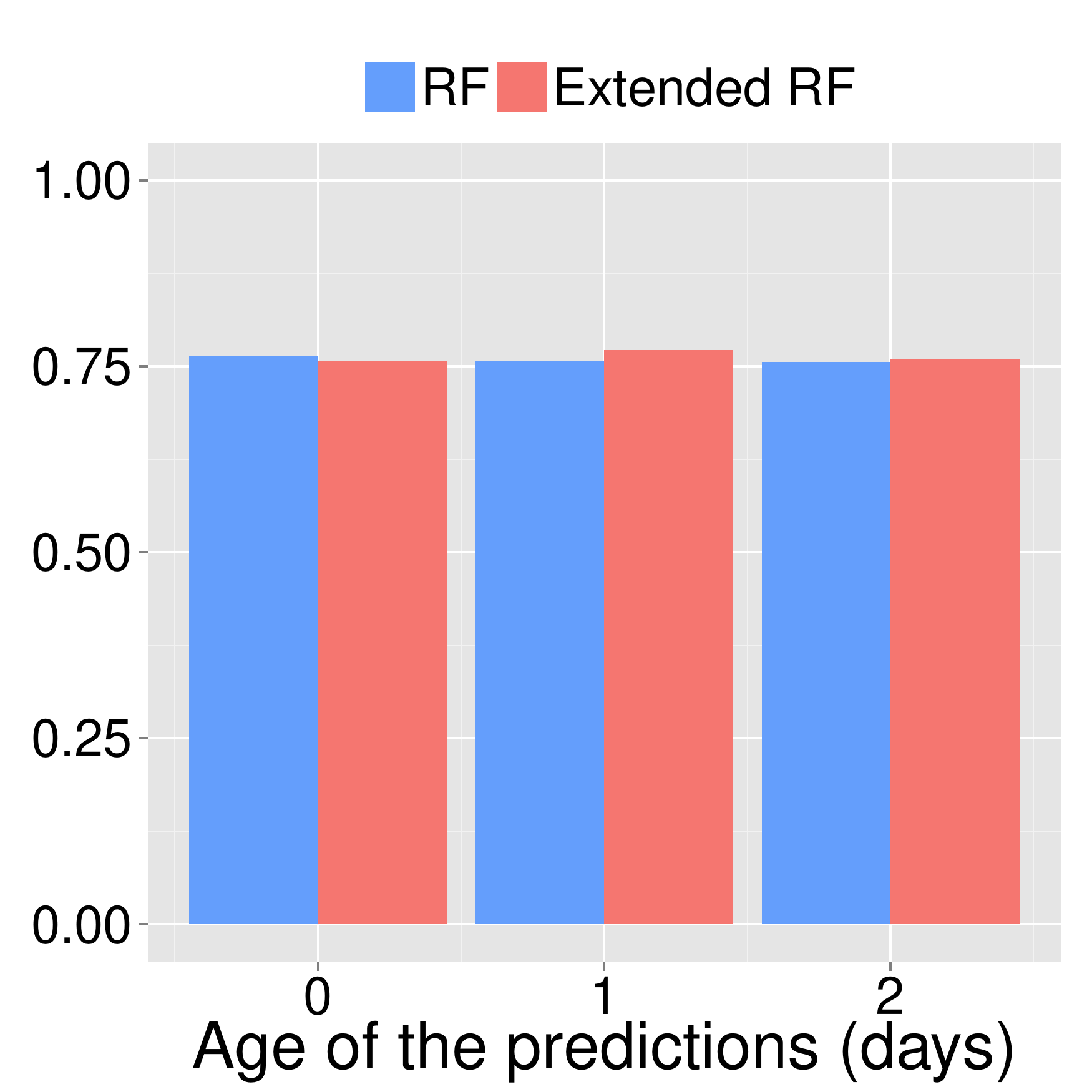}
                \caption{Specificity}
                \label{fig:rf-consolided-Specificity}
        \end{subfigure}%
        \caption{Performance of the predictions according to their age.}
        \label{fig:rf-consolided}
\end{figure*}

The results (illustrated in Figure~\ref{fig:rf-consolided}) show that the accuracy 
of the predictions using only the data 
from the \emph{Bicing} system decreased from $49.1\%$ to $47\%$, which
may not be considered a big difference, but the most complex models performed always
better and kept their accuracy over $49.4\%$ even when doing predictions for
$2$ days later. 
Moreover, the predictions that used external information not only had better 
sensitivity when done some days before, but also performed better than the 
simplest models.
For example, the sensitivity of the extended models is around $43.2\%$ for 
the predictions about the next $24$ hours and $48.9\%$ when about the next $48 - 
72$ hours. The specificity also increased $0.6\%$, which we do not consider a 
significant change, but it shows that the quality of the predictions can be 
maintained (if not improved) by the use of external information.

We conclude that the weather forecast complemented our models with information about 
the next days, which was useful enough to improve the quality of the predictions when 
we had no other information about the city's environment in the future.

\section{Conclusion and Future Work}
\label{section:conclusion}

In this work we presented an analysis of the data about the public bicycle 
system in Barcelona and the predictions about the statuses of the stations during 
$8$ days (February 5--12, 2015).
In order to run tests and draw our conclusions, we chose $4$ stations which  
often have a critical status, i.e., on average, they spend more than $5$ hours per 
day completely empty or completely full of bikes.

Using public data, we first observed the characteristics of such stations, for 
example, the impact of high temperatures and the influence of the relative 
humidity in the number of bicycles. 
Later, we compared the accuracy of the predictions using ARIMA with those using 
the Random Forest algorithm.
It was shown that their predictions using ARIMA models are inaccurate when the
goal is to inform about the critical states (i.e., \emph{full} and \emph{empty} 
stations), and cannot provide reliable information for the users.
Moreover, although ARIMA models have been shown to be a good option in the short-term 
(less than one hour), they do not perform well for longer periods, besides 
having higher complexity and consuming more computational resources to be 
created.

The predictions using the Random Forest algorithm performed better and were
able to correctly predict nearly half of the times when the stations were either
completely full or completely empty, up to $2$ days before they actually happened.
Also, the sensitivity of the predictions that use the Random Forest
algorithm is about $60\%$, which means that most of the times the "positive" outcomes
are correct. Furthermore, their specificity is around $75\%$, which means that 
every $4$ times that the models predict that there will bikes and free slots at a station, $3$
of them are correct.
We remark, nonetheless, that they may require some improvement before being 
adopted by applications aimed at users of the bicycle system.
The use of other relevant predictors may be an option to build more powerful 
models and can be done by using more observations and data from other sources.
For example, non-public information about the times when the bikes will be 
collected or taken to a station should have a positive impact in their performance.


From the results shown in this paper, it is possible to observe that the use of 
the bike stations is partially predictable and that, based on the predictions 
done using only the data accessible by the public, it would be possible to 
improve the support schedule.
In case of taking actions to improve the service, the system is expected to
evolve and new trends may be observed. 
However, given that the models may be regenerated every $3$ days, they are able to
incorporate such novelties, as well as variant user behaviors across different
times of the year.

From the point of view of the system administration, the predictions may 
trigger different actions, such as collecting bikes from a station that is 
almost full before the users face the problem of not finding places to leave 
their bikes. 
Moreover, it is possible to extend these predictions with other datasets 
available online, like the neighborhood wealth, the proximity of a place to 
other public transportation, schools and companies. 
The extended version of the predictions may be used to decide whether it is a 
good option to install a new station at a certain place or not, based on how 
many users would use it during the year.

From the user's perspective, this set of predictions may be used as a framework 
to improve their current API and show not only the current status of the 
stations, but also the future statuses.
Our future plans include organizing the predictions for all stations in a 
scalable way and provide to the users an interface to access this information. 
The interface can be a mobile application that provides users the option to 
make plans based on the predictions about the number of bikes at the selected 
stations.

\section*{Acknowledgment}

This work has been partially supported by the Spanish Government through the 
project TEC2012-32354 (Plan Nacional I+D) and by the Catalan Government 
through the project SGR-2014-1173.

\bibliographystyle{IEEEtran}

\bibliography{IEEEabrv,bibliography}
\end{document}